# A Novel Hybrid Algorithm for Optimized Solutions in Ocean Renewable Energy Industry: Enhancing Power Take-Off Parameters and Site Selection Procedure of Wave Energy Converters

Hossein Mehdipour[a], Erfan Amini[b], Seyed Taghi (Omid) Naeeni[a], Mehdi Neshat[c,d]

[a]School of Civil Engineering, College of Engineering, University of Tehran, Tehran 14179, Iran
hossein.mehdipour@ut.ac.ir, stnaeeni@ut.ac.ir

[b]Department of Civil, Environmental and Ocean Engineering, Stevens Institute of Technology, Hoboken, New Jersey 07030, United States
eamini@stevens.edu

[c]Center for Artificial Intelligence Research and Optimization, Torrens University Australia, Brisbane, QLD 4006, Australia
mehdi.neshat@torrens.edu.au

[d]Faculty of Engineering and Information Technology, University of Technology Sydney, Ultimo, Sydney, 2007, NSW, Australia
mehdi.neshat@uts.edu.au

Abstract

Ocean renewable energy, particularly wave energy, has emerged as a pivotal component for diversifying the global energy portfolio, reducing dependence on fossil fuels, and mitigating climate change impacts. This study delves into the optimization of power take-off (PTO) parameters and the site selection process for an offshore oscillating surge wave energy converter (OSWEC). The intricate interplay between waves and the energy conversion device results in nonlinear and transient responses in wave energy converters (WECs). Hence, there's a pressing need to optimize PTO parameters, ensuring that the device maximizes power absorption while evading potential damage or instability. However, the intrinsic dynamics of these interactions, coupled with the multi-modal nature of the optimization landscape, make this a daunting challenge. Addressing this, we introduce the novel Hill Climb - Explorative Gray Wolf Optimizer (HC-EGWO). This new methodology blends a local search method with a global optimizer, incorporating dynamic control over exploration and exploitation rates. This balance paves the way for an enhanced exploration of the solution space, ensuring the identification of superior-quality solutions. Further anchoring our approach, a feasibility landscape analysis based on linear water wave theory assumptions and the flap's maximum angular motion is conducted. This ensures the optimized OSWEC consistently operates within safety and efficiency parameters. Our findings hold significant promise for the development of more streamlined OSWEC power take-off systems. They provide insights for selecting the prime offshore site, optimizing power output, and bolstering the overall adoption of ocean renewable energy sources. Impressively, by employing the HC-EGWO method, we achieved an upswing of up to 3.31% in power output compared to other methods. This substantial increment underscores the efficacy of our proposed optimization approach. Conclusively, the outcomes offer invaluable knowledge for deploying OSWECs in the South Caspian Sea, where unique environmental conditions intersect with considerable energy potential.

1. Introduction

The importance of ocean renewable energy cannot be overstated, as it offers a promising means to diversify the global energy portfolio, reduce dependence on fossil fuels, and mitigate the impacts of climate change [1]. Due to the vastness and untapped potential of the world's oceans, harnessing their power for sustainable electricity generation is critical for meeting the rising energy demands of an ever-growing global population. Furthermore, ocean energy resources such as tidal, wave, and ocean thermal energy conversion (OTEC) exhibit lower variability and higher predictability compared to other renewable sources like wind and solar [2]. However, it is the ocean wave energy sector that has witnessed the most substantial advancements in recent years, with numerous ocean wave energy converters (WECs) under development and testing [3, 4]. These devices capture and transform the kinetic and potential energy present in ocean waves into electricity [5]. Moreover, their environmental and economic feasibility have also been investigated.

There are several methods for WEC classification. The first one is based on location. The WEC can be located at the shoreline or offshore. Offshore devices can harvest greater amounts of energy. The following criterion is how the device operates; it can be divided into submerged pressure differential, oscillating water column, overtopping device, or oscillating surge wave energy converter [6], which is the most popular one [7]. In this research, an offshore OSWEC device is investigated.

The vigorous surge motion, cost-effective installation, and minimal environmental impact have made OSWECs a preferable choice[8]. Numerous studies have investigated the potential of Oscillating Surge Wave Energy Converters (OSWECs) as a viable wave energy conversion technology; for instance, Ghasemipour et al. inspected nearshore regions of the southern coast of Iran for the feasibility of such devices [9]. Folley et al. have studied the effects of water depth [10] and device width [11] on the performance of OSWECs. The effects of the device's flap's width [12], length [13], orientation [14], shape, weight, and thickness [15] on the converter's performance have also been studied. It's been shown that the increase in the OSWEC's PTO has positive effects on power and flap's motion amplitude up to a certain point [16]. The wave characteristics like frequency [11] and period [17] can also influence the OSWEC's performance. Moreover, Lin et al. showed that, on average, the viscous loss of the fluid decreases the capture factor by 20% [18].

Almost all the numerical simulations in recent years have been based on Computational Fluid Dynamics (CFD). On one end of the spectrum of these methods is the Linear Potential Flow theory models [19], which are fast but not very accurate. On the other hand, some studies [20, 21] have used Reynolds Averaged Navier–Stokes Equations (RANS) CFD solvers for WEC analysis and simulation, which are computationally complex and slow but offer higher fidelity [22]. Recently, the WEC-Sim module, designed for MATLAB and LPF-based, has been extensively used for WEC simulations [23]. Different types of converters like point absorbers [24], OSWECs [25, 26, 27, 28], FOSWECs [29, 30, 31], and even novel WEC types [32] have been inspected using WEC-Sim. WEC-Sim is an open-source simulation tool designed for WEC numerical simulations [33]. Much research has been utilizing WEC-Sim to investigate OSWEC performance, which encompasses a variety of objectives, for instance, minimizing cost [25], reducing the hydrodynamic loads [34], lowering the applied loads to the support structure of the device [26], and mitigating the horizontal motion of the OSWEC's platform which in turn reduces costs [31]. In the early studies of wave energy converters, the predominant focus of numerical studies was on linear PTOs. For instance, Sheng et al. have optimized two models of linear PTOs for a Wave-Activated Bodies WEC [35] and an OWC [36]. However, researchers have since shown interest in the performance analysis of WECs with a Hydraulic PTO [37, 38, 39].

A variety of optimization studies have also been used in this field. In [40], an improved version of the differential evolution (DE) algorithm was used for a WEC array, simultaneously achieving more precise convergence and speed. Gomes et al. [41] did a hull optimization of a floating OWC using DE and COBYLA, a direct search method to achieve maximum power output. The Genetic Algorithm (GA) has been used widely in the field of wave energy generation; for instance, in [42], it has been used for shape optimization of a planar pressure differential WEC, and in [43], the WEC array configuration was optimized as well.

In [44], multiple meta-heuristic optimization algorithms, like GA or Particle Swarm Optimization (PSO), were used for the geometry optimization of WECs. The PSO has also been used for the optimization of WEC systems [45, 46]. Furthermore, Neshat et al. devised the improved Moth-Flame Optimizer (MFO) to optimize the geometry and the PTO settings of a generic multi-mode WEC [47]. Moreover, the GWO has been utilized for optimization in the field of other sources of renewable energy as well [48, 49].

This study proposes a fast and effective hybrid optimization method for maximizing the power absorption of an OSWEC based on the hindcast wave data from nine zones in the Caspian Sea, each has 9-12 data points. The significant contributions of this work are as follows:

- Proposing a novel optimizer to maximize the power absorption of an OSWEC, the Hill Climb-Explorative Gray Wolf Optimizer (HC-EGWO) methodology combines a local search method with a global optimizer to balance exploration and exploitation rates for improved solution quality.

- Developing a technical feasibility landscape analysis utilizing the Wave Energy Converter Simulator (WECSim) numerical model to account for the maximum feasible angular motion of the flap, ensuring optimized OSWEC operation within safety and efficiency limits.

- Insights for selecting optimal offshore sites, optimizing power output, and promoting the adoption of ocean renewable energy sources.

- Achieving a significant increase in power output (up to 3.31%) compared to other methods demonstrates the effectiveness of the proposed HC-EGWO optimization approach.

- Gaining valuable knowledge for deploying OSWECs in the South Caspian Sea, considering its unique environmental conditions and energy potential.

This study is organized as follows. In section 2, the data collection, WEC's feasibility, and other preliminary analyses are presented. Section 3 goes over the multiple modifications of GWO and proposes a new optimization algorithm. The following section provides the benchmark functions used to evaluate the new algorithm's performance. Section 5 presents the problem formulation information. Finally, Section 6 provides the results of the study.

## 2. Case Study Landscape Analysis

In this study, both the analytical model and the GWO algorithm were utilized for numerical modeling. The equations in each section were formulated and implemented in MATLAB. Subsequently, the solutions obtained from the proposed optimization approach were duly validated.

### 2.1. The Caspian Sea

The Caspian Sea is between Iran to the south, Russia to the north, Russia and Azerbaijan to the west, and Turkmenistan and Kazakhstan to the east. This body of water is often categorized either as the largest lake in the world or the most miniature sea on Earth. It spans approximately 1030-1200 kilometers in length and 196-435 kilometers in width. The surface of the Caspian Sea lies around 28 meters below sea level. The northern part of the sea is notably shallow, with only a negligible portion of seawater present in the northern quarter and an average depth of less than 5 meters [50]. Hence, investigating the southern shores becomes more significant for wave analysis. Figure 1 provides an overview of the Caspian Sea. Due to its status as one of Asia's most crucial energy sources, the Caspian Sea has always attracted considerable attention from the industry. The expansion of its southern coast also presents significant potential for harnessing wave energy [51, 52].

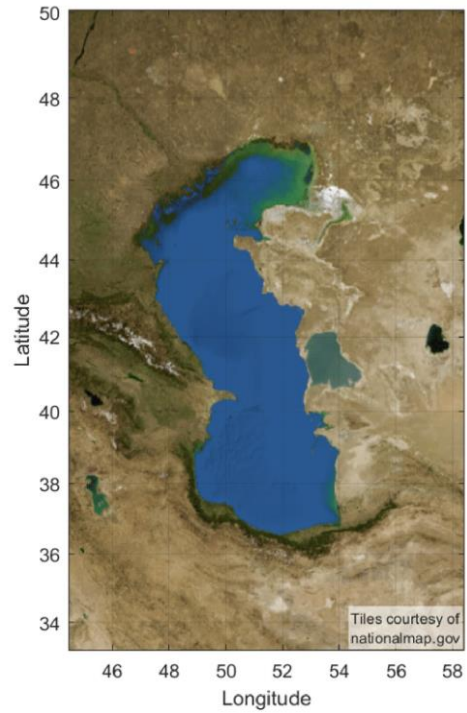

Figure 1: Caspian Sea landscape [50]

Various analyses have been conducted to forecast waves in the southern areas of the Caspian Sea, considering the prevailing direction of the dominant sea waves. Figure 2 displays the projected values of 50-year dominant waves in the southern regions, utilizing the Gumble distribution. Given the wave heights depicted in Figure 2, it becomes crucial to identify a point with maximum wave energy that also offers convenient beach access. Therefore, comprehensive research is needed to analyze wave data in the southern Caspian Sea, aiming to identify this point and establish a general criterion for comparing energy levels among different points using parameters such as wave height and wave period.

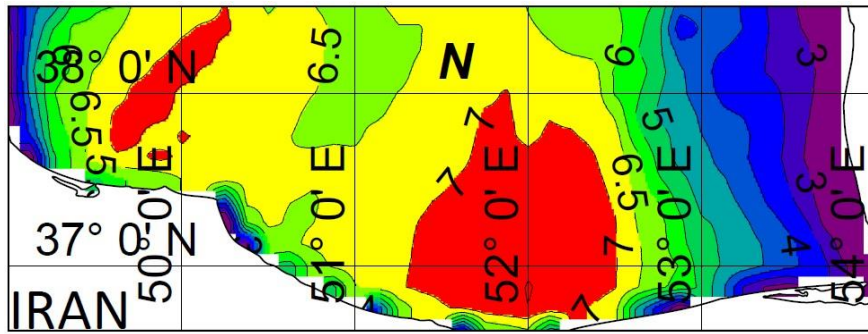

Figure 2: the 50-year design wave height in the dominant directions of the southern Caspian Sea [53]

*2.2. Data Collection*

To investigate the southern coasts of the Caspian Sea, the initial step involved analyzing the data obtained from the Iranian National Institute for Oceanography and Atmospheric Science (INIO). Specifically, the applied data from the Iranian Seas Wave Modeling (ISWM) and the Iranian Wave Atlas (IWA) models were examined. These datasets covered the entire Caspian Sea over a five-year period, from January 2006 to December 2010, with 1-hour time intervals. With reference to relevant literature and local assessments, nine ports were selected on the southern coasts of the Caspian Sea. A designated area of 0.2 longitude and latitude was considered around these ports, and locations with available data within this area were extracted. In total, 105 data points from the southern coast area of the Caspian Sea were identified, and their specifications are detailed in Figure 3 and Figure 4.

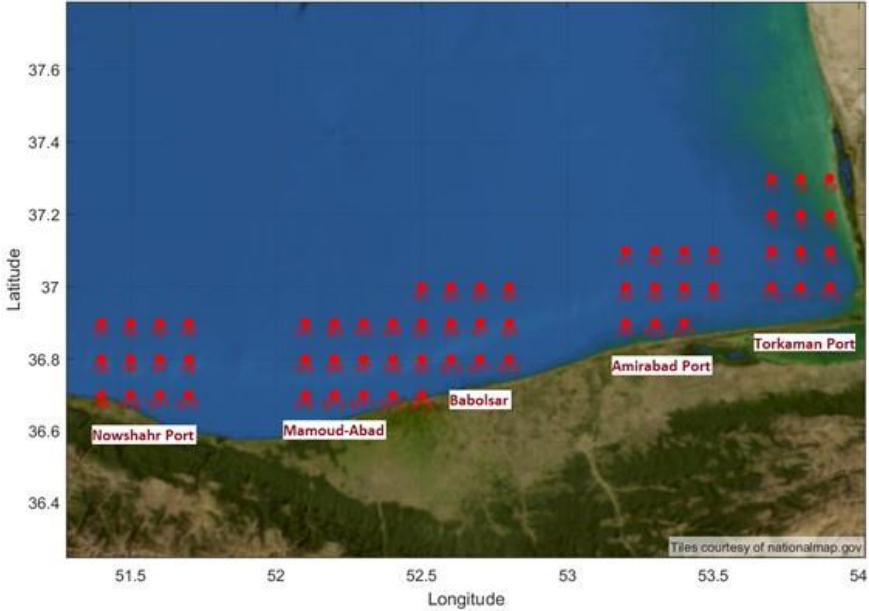

Figure 3: A representation of the surveyed areas along the southeastern coast of the Caspian Sea, including Torkaman, Amirabad, Babolsar, Mahmoud-Abad, and Nowshahr Port.

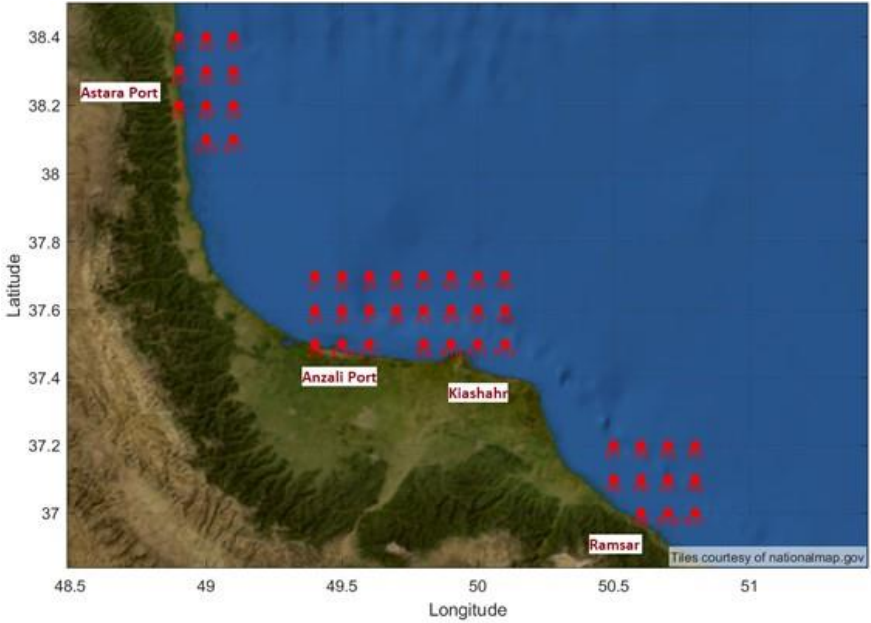

Figure 4: A representation of the surveyed areas along the southeastern coast of the Caspian Sea, including Ramsar, Kiashahr, Anzali Port, and Astara Port.

*2.3. Preliminary Analysis*

In order to understand the waves in the Caspian Sea, the data from nine selected ports were visualized. This was done by plotting wave rose diagrams (Figure 5) and wave scatter diagrams (Figure 6) to visualize the distribution of wave directions and to identify the prevailing wave patterns in the region. The variations in wave height and wave period across different locations in the study area were unveiled by analyzing the wave scatter diagrams. As seen in Figure 5, the waves have a relatively small magnitude and are mainly to the north, which is logical because these ports are in the southern part of the Caspian Sea. Moreover, from the scatter wave diagram in Figure 6, the waves comparably have low heights and low periods, and the most prevalent waves have a height of 20 cm and a period of 3 seconds.

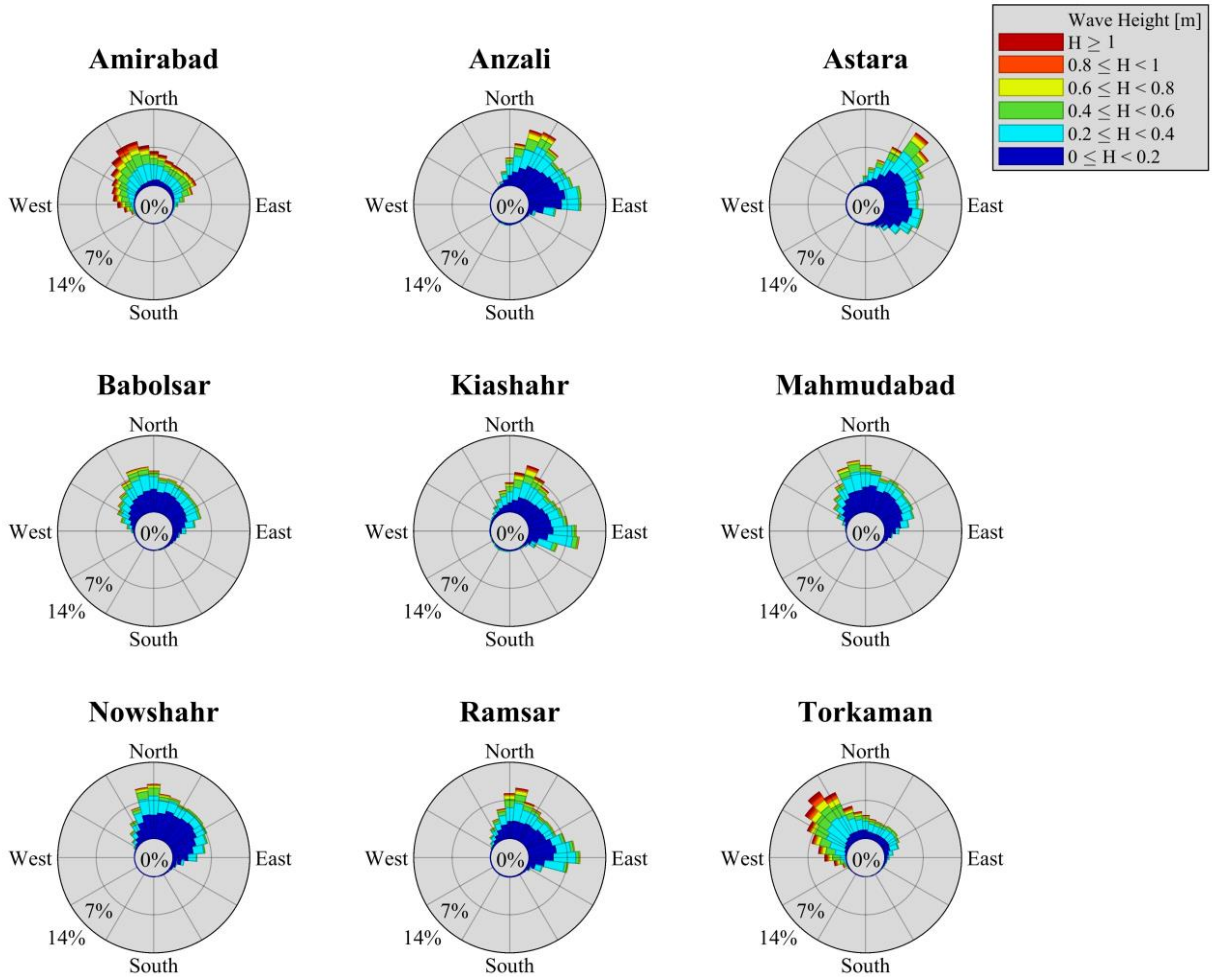

Figure 5: the wave rose diagram for the nine analyzed sea ports reveals relatively small wave magnitudes that predominantly originate from the north. This observation aligns with expectations since these ports are located in the southern region of the Caspian Sea.

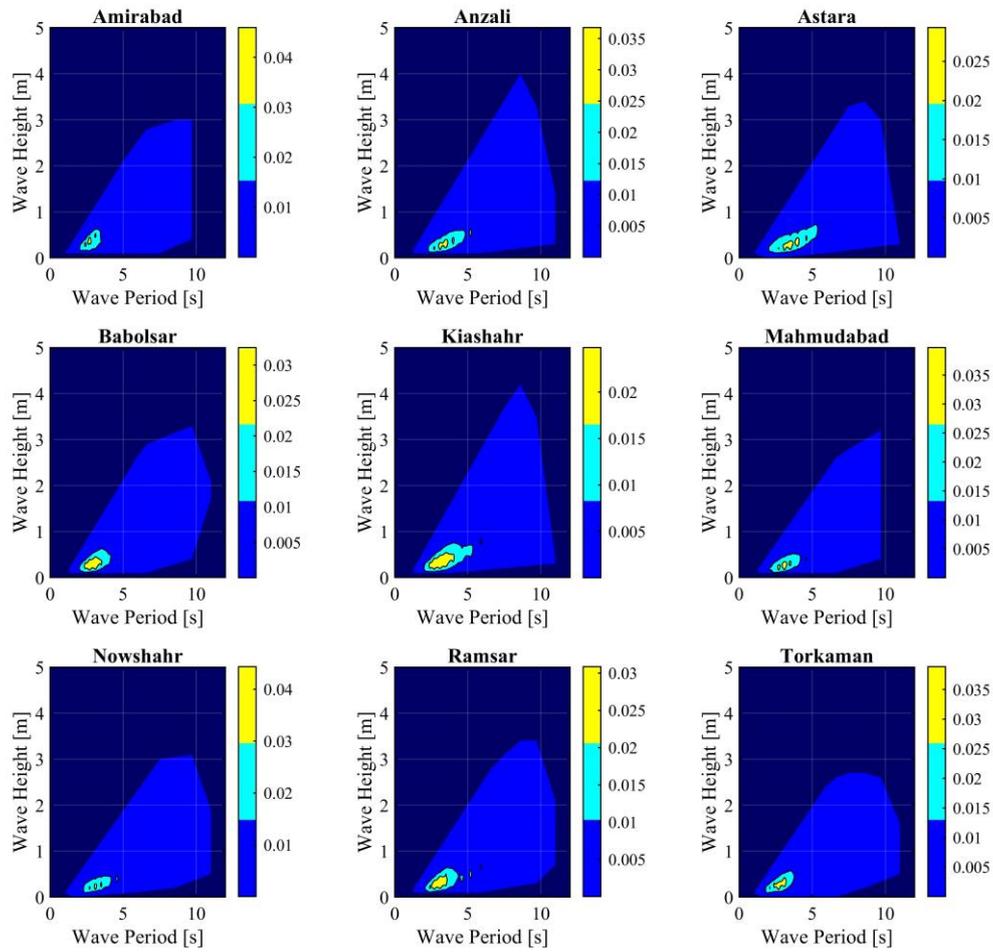

Figure 6: The wave scatter diagram of the nine studied ports.

A power matrix was also made specifically for the Caspian Sea, shown in Figure 7. This matrix comprehensively assessed the wave energy potential in these regions. It takes into account important factors such as wave height and wave period to estimate the energy conversion capabilities of OSWECs in these areas. As shown in Figure 7, increasing either the height or the period of the wave can lead to higher absorbed energy.

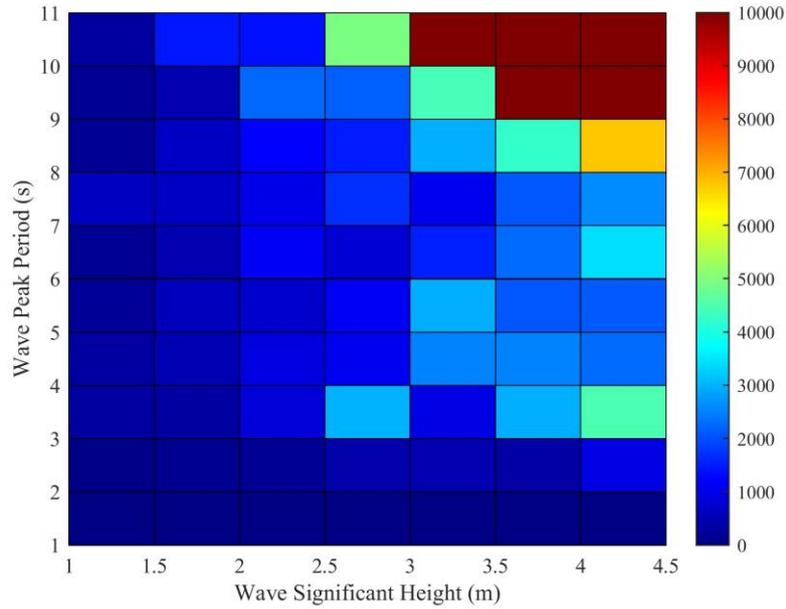

Figure 7: The power matrix of the southern coasts of the Caspian Sea.

Collectively, the insights gained from the wave rose diagrams, wave scatter diagrams, and power matrix contribute to our understanding of the spatial distribution of wave energy in the Caspian Sea.

*2.4. WEC's Feasibility*

Analysis of OSWEC's flap interaction with the wave under linear water wave theory assumptions requires the flap's excursions to be adequately small. The reason is that the flap's rotation should be small enough so that the correct and non-linear form of the hydrostatic stiffness, which is ($K_p \sin\theta$), can be replaced by ($K_p$) [54]. Several studies [55, 17, 56, 16, 57] assumed the maximum angular motion of the flap to be 30 °) [55]. In addition, this limitation helps the device to avoid damage, particularly in extreme sea states [56]. In Figure 8, the feasible area of the damping and stiffness of the PTO are presented. First, the literature for the range of viable damping and stiffness values for the PTO was reviewed, the result of which is shown by the orange color. Next, based on the OSWEC's flap oscillation limitation, the feasible values of $K_{PTO}$ and $C_{PTO}$ were finalized, and the red area was omitted. Finally, the remaining area, shown by the color green, represents the applicable range of these two critical parameters.

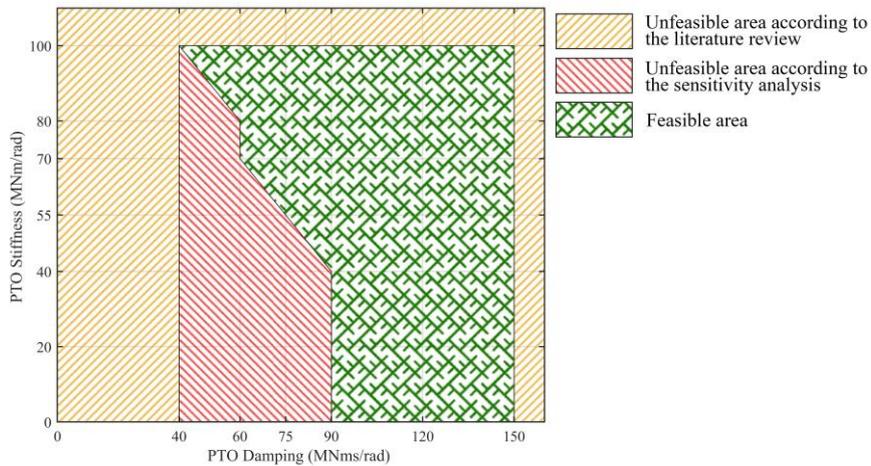

Figure 8: The feasible area of the PTO damping and stiffness used in this study.

*2.5. Preliminary Sensitivity Analysis*

In order to investigate the impact of critical parameters on the performance of an oscillating wave energy converter, a sensitivity analysis was conducted. Wave height (H), wave period (T), PTO's damping (C), and stiffness (K) were analyzed regarding the power output of the system through six plots in Figure 9. By examining the plots, valuable insights were obtained regarding the optimal values for these parameters and their combinations for maximizing power generation.

Figure 9-(a) represents the effect of H and T in optimizing the generated power of the OSWEC. The plot reveals that a combination of high wave height and wave period leads to the best power outputs. However, it is noteworthy that extreme values of T can decrease the power. Figure 9-(b) illustrates the influence of K and H on the converter's power generation. As can be seen, both high and low levels of PTO stiffness can result in passable power outputs. However, the highest power is achieved when K is moderate. Figure 9-(c) depicts the relation between C and H and the device's generated power. High wave heights and low values of PTO damping correspond to favorable power outputs and improved performance.

Figure 9-(d) shows the effects of K and T variation on power. Accordingly, the wave periods within the range of approximately 6 to 9 seconds yield pleasing power outputs. Furthermore, the power improves significantly as the PTO stiffness approaches its medium value. Figure 9-(e) showcases the influence of C and T. Similar to the previous plot, wave periods ranging from approximately 6 to 9 seconds produce the most favorable power outputs— additionally, the performance improves as the PTO damping decreases. Finally, in Figure 9-(f), the PTO parameters regarding their effect on the power output have been investigated. According to the plot, almost always lower values C result in better power, and a K value between 10 and 70 MNm/rad leads to the best performance. Overall, the parameter's effects can be explained relatively simply. In summary, higher H and lower C lead to the best power outputs. Furthermore, moderate values of T and K lead to the best performance. Overall, it can be seen that the problem at hand is a multimodal optimization problem and has multiple optima.

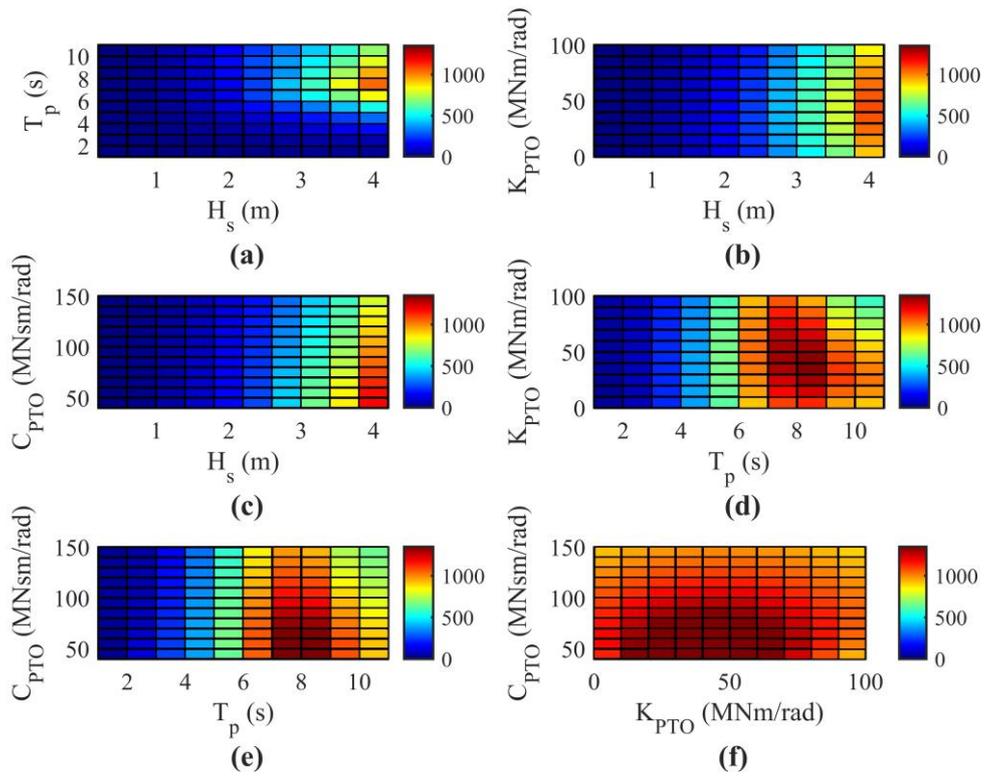

Figure 9: Sensitivity analysis plots for the key parameters of the OSWEC (Wave height, wave period, PTO damping, and PTO stiffness)

Next, the effects of the PTO parameters on the PTO power were further inspected in Figure 10. The black parts show the unfeasible areas calculated in previous sections. Similar to the power output, low values of C and moderate K values bring about the highest PTO forces. It is worth noting that in the OSWEC, the power output is calculated by multiplying PTO force by the flap's velocity [17].

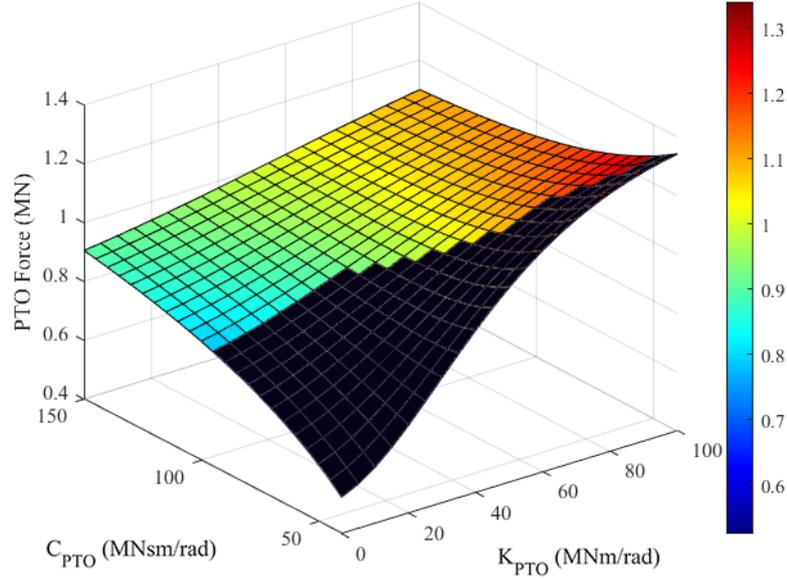

Figure 10: Sensitivity analysis result of the effects of PTO parameters on the PTO force. Unfeasible values, shown in black, are excluded from the analysis.

## 3. Optimization Approach

### 3.1. The Standard GWO

The Grey Wolf Optimizer (GWO) is a bio-inspired algorithm based on a grey wolf breed's leadership hierarchy and hunting behaviour. Mirjalili et al. [58] simplified their hunting mechanism and introduced four types of wolves, the alpha (α), the beta (β), the delta (δ), which are, respectively, the best solutions of the algorithms (have the best knowledge about the location of the optimum) and the omegas (ω) which comprise the rest of the pack and follow the three aforementioned wolves to get closer to the prey.

We can show the encircling of the prey process mathematically using the following equations:

$$\vec{D} = |\vec{C} \cdot \vec{X_p}(t) - \vec{X}(t)| \tag{1}$$

$$\vec{X}(t+1) = \vec{X_p}(t) - \vec{A} \cdot \vec{D} \tag{2}$$

in which $t$ is the current iteration, $\vec{X}$ indicates the position of a grey wolf, $\vec{X_p}$ is the position of the prey, and $\vec{A}$ and $\vec{C}$ are coefficient vectors. The $\vec{A}$ and $\vec{C}$ vectors are determined as follows:

$$\vec{A} = 2\vec{a} \cdot \vec{r_1} - \vec{a} \tag{3}$$

$$\vec{C} = 2 \cdot \vec{r_2} \tag{4}$$

in which $\vec{a}$ linearly decreases from 2 to 0, and $\vec{r_1}$ and $\vec{r_2}$ are random numbers between 0 and 1. As stated, the $\omega$ wolves update their positions based on the three best search agents ($\alpha, \beta$, and $\delta$ wolves). They follow these equations:

$$\vec{D_\alpha} = |\vec{C_1} \cdot \vec{X_\alpha} - \vec{X}| \tag{5}$$

$$\vec{D_\beta} = |\vec{C_2} \cdot \vec{X_\beta} - \vec{X}| \tag{6}$$

$$\vec{D_\delta} = |\vec{C_3} \cdot \vec{X_\delta} - \vec{X}| \tag{7}$$

$$\vec{X_1} = \vec{X_\alpha} - \vec{A_1} \cdot (\vec{D_\alpha}) \tag{8}$$

$$\vec{X_2} = \vec{X_\beta} - \vec{A_2} \cdot (\vec{D_\beta}) \tag{9}$$

$$\vec{X_3} = \vec{X_\delta} - \vec{A_3} \cdot (\vec{D_\delta}) \tag{10}$$

$$\vec{X}(t+1) = \frac{\vec{X_1}(t) + \vec{X_2}(t) + \vec{X_3}(t)}{3} \tag{11}$$

This was a simple overview of GWO's origin and mechanism.

### 3.2. Modified GWO (mGWO)

Mittal et al. [59] believed that the linear equation of the a does not provide a good balance between exploration and exploitation, so they tried this nonlinear equation:

$$\vec{a} = 2\left(1 - \frac{t^2}{T^2}\right) \tag{12}$$

in which $t$ represents the current iteration, and $T$ is the total number of iterations. This equation resulted in 70% exploration and 30% exploitation of the total iterations.

### 3.3. Exploration-Enhanced GWO (EEGWO)

Since in GWO, all the search agents gravitate toward the three best solutions, this algorithm can be susceptible to premature convergence. Therefore, Long et al. [60] modified the position-updating equation inspired by the PSO algorithm to emphasize more on the exploration:

$$\vec{X}(t+1) = b_1 \cdot r_3 \cdot \frac{\vec{X_1}(t) + \vec{X_2}(t) + \vec{X_3}(t)}{3} + b_2 \cdot r_4 \cdot (\vec{X'} - \vec{X}) \tag{13}$$

where $\vec{X'}$ is another randomly selected search agent from the population, $r_3$ and $r_4$ are random numbers in [0,1], and $b_1, b_2 \in (0,1)$ indicate constant coefficients to balance the exploration/exploitation (in the mentioned study the selected values are $b_1 = 0.1$ and $b_2 = 0.9$).
They also proposed a new formula for the control parameter $\vec{a}$:

$$\vec{a} = a_{\text{initial}} - (a_{\text{initial}} - a_{\text{final}}) \cdot \left(\frac{T-t}{T}\right)^\mu \tag{14}$$

where $\mu$ is the nonlinear modulation index ( $\mu = 1.5$ in the aforementioned study), and $a_{\text{initial}}$ and $a_{\text{final}}$ are 2 and 0, respectively.

### 3.4. Improved GWO (IGWO)

In order to address the challenges associated with the conventional Maximum Power Point Tracking (MPPT) techniques, which is the power maximization of the PV system [61], and improving their efficiency in finding the global maximum power point, Ma et al. [62] utilized the fitness value of the search agents for their position-updating mechanism as follow:

$$\vec{X}(t+1) = \begin{cases} \frac{f_\alpha \cdot \vec{X_1}}{f} + \frac{f_\beta \cdot \vec{X_2}}{f} + \frac{f_\delta \cdot \vec{X_3}}{f}, \\ \frac{\vec{X_1} + \vec{X_2} + \vec{X_3}}{3}, \end{cases} \tag{15}$$

$$f = f_\alpha + f_\beta + f_\delta \tag{16}$$

where $f_\alpha, f_\beta$, and $f_\delta$ are fitness values of $\alpha, \beta$, and $\delta$, respectively. $f_{\text{avg}}$ is the average of these 3 fitness values, and $f_i$ is the fitness value of grey wolf individuals.

The authors modified the $\vec{a}$ formula as well:

$$\vec{a} = a_{\min} + (a_{\max} - a_{\min}) \cdot \left(1 - \frac{t}{T}\right)^2 \tag{17}$$

where $a_{min}$ and $a_{max}$ are 0 and 2, respectively.

### 3.5. Efficient and Robust GWO (ERGWO)

With the intention of tackling large-scale numerical optimization problems, Long et al. performed another study to enhance the performance of the GWO [63]. Following the footsteps of the previous studies, they changed both the position-updating equation and a equation. The first change can be seen below, where they used a proportional weighting method similar to [62]:

$$w_1 = \frac{|\vec{X_1}|}{|\vec{X_1}| + |\vec{X_2}| + |\vec{X_3}|} \tag{18}$$

$$w_2 = \frac{|\vec{X_2}|}{|\vec{X_1}| + |\vec{X_2}| + |\vec{X_3}|} \tag{19}$$

$$w_3 = \frac{|\vec{X_3}|}{|\vec{X_1}| + |\vec{X_2}| + |\vec{X_3}|} \tag{20}$$

$$\vec{X}(t+1) = \frac{1}{w_1 + w_2 + w_3} \cdot \frac{w_1 \cdot \vec{X_1} + w_2 \cdot \vec{X_2} + w_3 \cdot \vec{X_3}}{3} \tag{21}$$

where $a_{initial}$ and $a_{final}$ are 2 and 0, respectively. $\mu \in [1.0001, 1.005]$ is the nonlinear modulation index (in the mentioned study $\mu = 1.001$).

$$\vec{a} = a_{\text{initial}} - (a_{\text{initial}} - a_{\text{final}}) \cdot \mu^{-t} \tag{22}$$

## 3.6. Hill-Climbing Exploraitive GWO (HC-EGWO)

The Gray Wolf Optimizer (GWO), although remarkable in its performance for solving diverse optimization problems [58], is not without its shortcomings. Primarily, it is prone to premature convergence towards local optima in the search space, specifically during complex, high-dimensional problems [64]. This challenge arises due to the declining exploration rate (parameter *a*) in the original GWO algorithm, which transitions from 2 to 0 linearly. While this allows the algorithm to either explore or exploit optimal solutions when *a* is above 1, it leads to exploitation when *a* is below 1, thereby accelerating convergence towards local optima.

In addressing this limitation, we propose the Explorative Gray Wolf Optimizer (EGWO), a novel enhancement to the original GWO that amplifies the exploration rate by modifying the parameter *a*. In EGWO, *a* is altered as per the equations:

$$R = \left(\frac{T-t}{T-1}\right) \tag{23}$$

$$\vec{a} = 2 \cdot \left(1 - \left(e^{(t^R - T^R)}\right)\right) \tag{24}$$

where *t* is the current iteration, and *T* is the maximum number of iterations. This modification empowers the algorithm to delay the convergence process and explore the search space more thoroughly, reducing the chance of being trapped in local optima. Furthermore, to fortify the global search capabilities of EGWO, we propose a robust hybrid algorithm that incorporates a Random-restart hill-climbing local search, dubbed HC-EGWO. Figure 11 shows the evolution of *a* throughout the iterations for each of the six optimization approaches. Also, the Exploration Ratio (ER) is presented for each method; this value shows how much of the search process is allocated to potential exploration in GWO. By comparing the values, it is clear that HC-EGWO has the best ER value and the most potential to search the unexplored areas of the search space thoroughly.

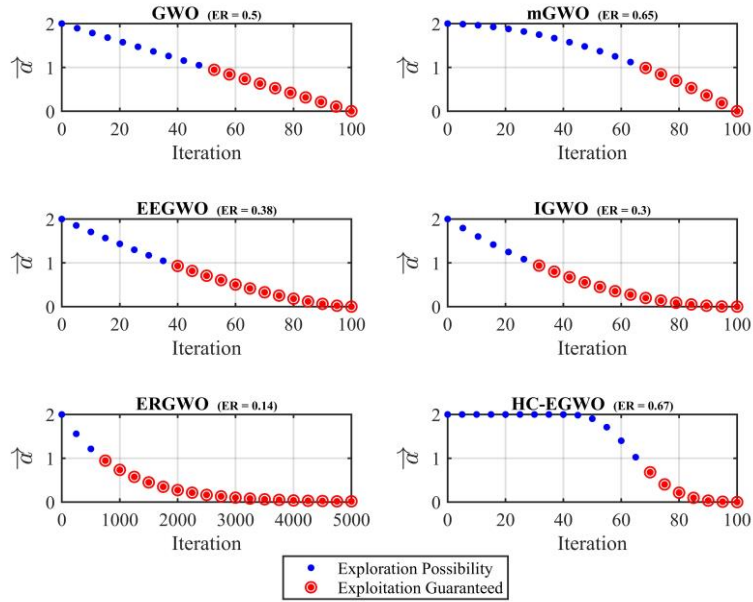

Figure 11: The evolution of *a* value during the optimization process for the six evaluated GWO methods in this study, and their Exploration Ratio (ER)

In this hybrid scheme, EGWO operates on a superior level to create a global track and procure an array of suitable solutions. When the EGWO encounters stagnation or converges prematurely towards a local optimum, the hill climbing algorithm initiates a local search around the best solution found by the upper level (EGWO). It does so by creating a comprehensive neighbourhood search, thereby preparing to escape such unfavourable scenarios.

The performance threshold is computed as follows:

$$\Delta \text{Best}_{THD} = \frac{\sum_{k=1}^{M}\left(\text{Best}_{THD_k} - \text{Best}_{THD_{k-1}}\right)}{M} \tag{25}$$

where $Best_{THD}$ is the optimal solution found per generation, and $M$ is tied to the range of iterations to determine the average EGWO performance. When the solution offered by the local search outperforms the initial one, EGWO's global best is updated.

The HC-EGWO algorithm iteratively executes the hill-climbing process whenever the EGWO performance dips, each time establishing an initial condition to facilitate escape from undesirable circumstances. The search step size is decremented linearly as follows to achieve a fine balance between exploration and exploitation:

$$S_t = S_t - \left(\frac{t}{T}S_t\right) + 1 \tag{26}$$

Here, $t$ and $T$ denote the current and maximum iteration numbers, respectively, while $S_t$ represents the neighborhood search's step size.

Algorithm 1 illustrates the detailed steps of the proposed optimization method (HC-EGWO). The initial solution encompasses wave height ($H$), wave period ($T$), PTO stiffness coefficient ($K$), and PTO damping coefficient ($C$).

One of the crucial parameters in the local search algorithm is $g$, which signifies the precision of the neighborhood search surrounding the globally optimal solution proposed by EGWO. A smaller step size for HC slows down the convergence speed. However, a larger step size bolsters the exploration capability, possibly at the expense of the exploitation capability, leading to the possibility of skipping over globally optimal or high-potential solution surfaces. For each decision variable, the neighborhood search evaluates two distinct direct searches, either incremental or decremental. After evaluating the generated solutions, the optimal candidate is chosen to iterate the search algorithm. It should be noted that the HC algorithm should not be employed during the initial iteration of the optimization process due to the pronounced tendency for converging to local optima. Moreover, for optimizing large-scale problems, HC may not be a suitable choice.

**Algorithm 1** *Hill Climbing Explorative Gray Wolf Optimizer*

1: **procedure** HC-EGWO
2:     $N = 30, D = 4$    ▷ Population size and dimension size
3:     $\mathbb{S} = \{\langle H_1, T_1, K_1, D_1\rangle, ..., \langle H_N, T_N, K_N, D_N\rangle\}$    ▷ Initialize the population of wolves
4:     Check if $lb_1^N \leq \mathbb{S} \leq ub_1^N$
5:     $Max_{iter} = 100$    ▷ Maximum number of iterations
6:     **for** $iter = 1, .., Max_{iter}$ **do**
7:        $R = (Max_{iter} - iter)/(Max_{iter} - 1)$
8:        $\vec{a} = 2 \cdot (1 - e^{(iter^R - Max_{iter}^R)})$    ▷ Calculate exploration rate $\vec{a}$ with the new formulation
9:        *Sort* the population $\mathbb{S}$ based on fitness and get the leading wolves $\alpha, \beta$, and $\delta$
10:        **for** $i = 1, .., N$ **do**
11:           **for** $j = 1, .., D$ **do**
12:              *Calculate* $A_{ij}$ and $C_{ij}$ for each of the leading wolves
13:              *Update* the position of the $i$th wolf in dimension $j$ using the positions of $\alpha, \beta$, and $\delta$
14:           **end for**
15:        **end for**
16:        *Update* the positions of $\alpha, \beta$, and $\delta$ based on the updated population
17:        $Best_{iter} = Max(\mathbb{S})$    ▷ Get the best solution in this iteration
18:        $\Delta Best = Best_{iter} - Best_{iter-1}$    ▷ Calculate the difference between the best solutions in the current and previous iterations
19:        **if** $\Delta Best < Th$ **then**    ▷ If the difference is less than a threshold $Th$, perform Hill Climbing
20:           *Initialize* the constraints $lb_1^d, ub_1^d$

```
21:         S_1^d = (Min_1^d + Max_1^d)/g                    ▷ Compute the step size, g is search resolution
22:         Sol_1 = {⟨H, T, K, D⟩}                           ▷ Initial solution
23:         (fitness_1)=Eval(Sol_1)                          ▷ Evaluate the solution
24:         for iter ≤ Max_iter do
25:             Te = Sol_iter
26:             while t ≤ len(Sol_1) do
27:                 Te_t = Te_t ± S_t                        ▷ Neighborhood search
28:                 (fitness_t^iter)=Eval(Te_t)
29:                 t = t + 1
30:             end while
31:             ⟨Max_fit, Index_max⟩=Max(fitness)
32:             Sol_iter=Te_t(Index_max)                     ▷ Select the best feasible solution and update the design
33:             S_t = S_t − (iter/Max_iter S_t) + 1           ▷ S_t linearly reduced
34:         end for
35:         Best_iter = Sol_iter
36:     end if
37: end for
38: return Best_iter                                         ▷ Return the best solution
39: end procedure
```

## 4. Benchmark Functions

In this section, we evaluate the performance of the HC-EGWO algorithm on a total of 16 benchmark functions. These are classical benchmark functions that have been widely used by researchers in the field. These functions are well-established and are commonly used to evaluate the performance of optimization algorithms. You can find a detailed list of these classical benchmark functions in Tables 1 and 2. The tables provide information such as the dimensionality (Dim) of the function, the range of the function's search space (Range), and the optimal value (fmin) of each function [58]. By benchmarking the HC-EGWO algorithm on these 16 functions, we can evaluate its performance and compare it to other optimization algorithms.

All the benchmark functions employed in this study are aimed at minimizing a given objective. These functions can be classified either as multimodal or fixed-dimension multimodal. To assess the performance of the HC-EGWO, it was executed 30 times on each benchmark function. The results were then analyzed statistically, providing the average and standard deviation values. These statistical outcomes are presented in Tables 5, which allow for comparing and evaluating the algorithm's performance across the benchmark functions. Also, a statistical analysis of the comparison is presented in the results section. To validate the results, the HC-EGWO algorithm is compared against other variations of the GWO algorithm, namely the conventional Grey Wolf Optimizer [58], the modified GWO [59], the Exploration-Enhanced GWO [60], the Improved GWO [62], and the Efficient and Robust GWO [63].

Table 1: Multimodal Benchmark Functions [58]

| Function | Dim | Range | $f_{min}$ |
|---|---|---|---|
| $F_1(x) = \sum_{i=1}^{n}(-x_i \cdot \sin(\sqrt{|x_i|}))$ | 30 | [-500, 500] | $-418.9829 \times 5$ |
| $F_2(x) = \sum_{i=1}^{n}(x_i^2 - 10 \cdot \cos(2\pi x_i) + 10)$ | 30 | [-5.12, 5.12] | 0 |
| $F_3(x) = -20 \cdot \exp\left(-0.2 \cdot \sqrt{\frac{\sum_{i=1}^{n} x_i^2}{n}}\right)$ $- \exp\left(\frac{1}{n}\sum_{i=1}^{n}\cos(2\pi x_i)\right) + 20 + e$ | 30 | [-32, 32] | 0 |
| $F_4(x) = \frac{1}{4000}\sum_{i=1}^{n} x_i^2 - \prod_{i=1}^{n}\cos\left(\frac{x_i}{\sqrt{i}}\right) + 1$ | 30 | [-600, 600] | 0 |
| $F_5 = \frac{\pi}{n}(10\sin(\pi y_1)) + \sum_{i=1}^{n-1}(y_i - 1)^2 \cdot$ $(1+10\sin^2(\pi y_{i+1}))+$ $(y_n - 1)^2 + \sum_{i=1}^{n} u(x_i, 10, 100, 4)$ $y_i = 1 + \frac{x_i+1}{4}$ $u(x_i, a, k, m) = \begin{cases} k(x_i - a)^m & x_i > a \\ 0 & -a < x_i < a \\ k(-x_i - a)^m & x_i < -a \end{cases}$ | 30 | [-100, 100] | 0 |
| $F_6(x) = 0.1\left(\sin^2(3\pi x_1)\right.$ $+ \sum_{i=1}^{n}\left((x_i - 1)^2 \cdot \left(1 + \sin^2(3\pi x_i + 1)\right)\right) +$ $(x_n - 1)^2 \left(1 + \sin^2(2\pi x_n)\right) + \sum_{i=1}^{n} U(x_i, 5, 100, 4)$ | 30 | [-50, 50] | 0 |

Table 2: Fixed-dimension Multimodal Benchmark Functions. [58]

| Function | Dim | Range | $f_{min}$ |
|---|---|---|---|
| $F_7 = \left(\frac{1}{500} + \sum_{j=1}^{25} \frac{1}{j+\sum_{i=1}^{2}(x_i-a_{ij})^6}\right)^{-1}$ | 2 | [-65, 65] | 1 |
| $F_8(x) = \sum_{i=1}^{11}\left(a_i - \frac{x_1(b_i^2+x_2 b_i)}{b_i^2+x_3 b_i+x_4}\right)^2$ | 4 | [-5, 5] | 0.00030 |
| $F_9(x) = 4x_1^2 - 2.1x_1^4 + \frac{1}{3}x_1^6 + x_1 x_2 - 4x_2^2 + 4x_2^4$ | 2 | [-5, 5] | -1.0316 |
| $F_{10}(x) = \left(x_2 - \frac{5.1}{4\pi^2}x_1^2 + \frac{5}{\pi}x_1 - 6\right)^2 + 10\left(1 - \frac{1}{8\pi}\right)\cos(x_1) + 10$ | 2 | [-5, 5] | 0.398 |
| $F_{11}(x) = \left(1 + (x_1 + x_2 + 1)^2(19 - 14x_1 + 3x_1^2 - 14x_2 + 6x_1 x_2 + 3x_2^2)\right)\left(30 + (2x_1 - 3x_2)^2(18 - 32x_1 + 12x_1^2 + 48x_2 - 36x_1 x_2 + 27x_2^2)\right)$ | 2 | [-2, 2] | 3 |
| $F_{12}(x) = -\sum_{i=1}^{4}\left(c_i \exp\left(-\sum_{j=1}^{3} a_{ij}(x_j - p_{ij})^2\right)\right)$ | 3 | [1, 3] | -3.86 |
| $F_{13}(x) = -\sum_{i=1}^{4}\left(c_i \exp\left(-\sum_{j=1}^{6} a_{ij}(x_j - p_{ij})^2\right)\right)$ | 6 | [0, 1] | -3.32 |
| $F_{14}(x) = -\sum_{i=1}^{5}\left(((X - a_i)(X - a_i)^T + c_i)^{-1}\right)$ | 4 | [0, 10] | -10.1532 |
| $F_{15}(x) = -\sum_{i=1}^{7}\left(((X - a_i)(X - a_i)^T + c_i)^{-1}\right)$ | 4 | [0, 10] | -10.4028 |
| $F_{16}(x) = -\sum_{i=1}^{10}\left(((X - a_i)(X - a_i)^T + c_i)^{-1}\right)$ | 4 | [0, 10] | 10.5363 |

## 5. Problem Formulation

### 5.1. The Wave Energy Converter

As stated before, the Oscillating Surge WEC was chosen for this study due to multiple reasons. The OSWEC is fixed to the ground, and it features a hinged connection between its base and flap. This hinge constrains the flap's movement, allowing it to pitch around the hinge point. The converter's physical dimensions at scale are shown in Figure 12. Moreover, the OSWEC's flap has a mass of 127 tonnes, and its other properties are listed in Table 3.

Table 3: OSWEC's Flap Mass Properties [23]

| Body | Direction | Center of Gravity (m) | $I_{xx}$ (kg.m$^2$) | $I_{yy}$ (kg.m$^2$) | $I_{zz}$ (kg.m$^2$) |
|---|---|---|---|---|---|
| Flap | x | 0 | 0 | 0 | 0 |
|  | y | 0 | 0 | 1,850,000 | 0 |
|  | z | -3.9 | 0 | 0 | 0 |

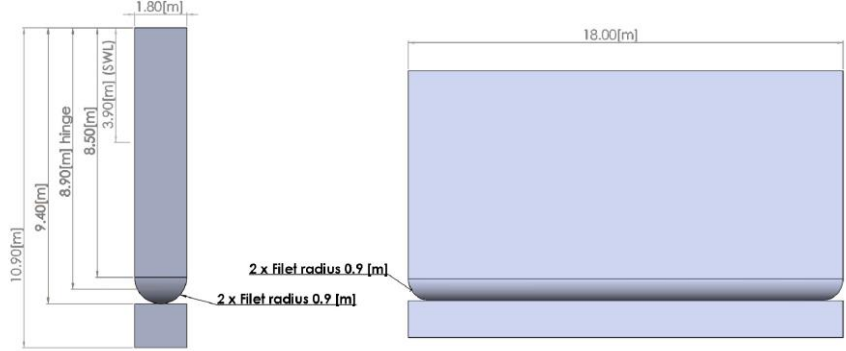

Figure 12: OSWEC's full-scale detailed dimensions [23]

## 5.2. WEC-Sim

WEC-Sim provides an open-source simulation tool for the community. In order to determine the dynamic response of the WEC system, the equation of motion for the device about its center of gravity in the time domain has to be solved [23]:

$$m\ddot{X} = F_{exc}(t) + F_{rad}(t) + F_{PTO}(t) + F_B(t) \qquad (27)$$

where $m$ is the mass matrix of the WEC, $\ddot{X}$ is the acceleration vector, $F_{exc}(t)$ is the wave excitation vector, $F_{rad}(t)$ is the force and torque vector caused by wave radiation, $F_{PTO}(t)$ is the PTO force and torque vector, and $F_B(t)$ is the net buoyancy restoring force and torque vector. The $F_{exc}(t)$ and $F_{rad}(t)$ are calculated using Boundary Element Method (BEM) solvers [65]. This module is developed on MATLAB/Simulink/Simscape. Figure 13 shows the Simulink models of the proposed OSWEC investigated in this paper [23]. Moreover, irregular waves are simulated as a superposition of regular waves [66].

In WEC-Sim, the PTO unit can be characterized by a linear spring-damper system, in which the PTO force is calculated by:

$$F_{PTO} = K \cdot X + C \cdot \dot{X} \qquad (28)$$

where K is the PTO stiffness coefficient, C is the PTO damping coefficient, and $X$ and $\dot{X}$ are the relative motion and velocity between the flap and the base of the OSWEC. Since the studied device is fixed to the bed, $X$ and $\dot{X}$ can be considered the flap's motion and velocity. Next, the power output of the PTO can be obtained by the following [23]:

$$P_{PTO} = F_{PTO} \cdot \dot{X} = K \cdot X \cdot \dot{X} + C \cdot \dot{X}^2 \qquad (29)$$

In WEC-Sim, the regular wave excitation force after the ramp time (the necessary time for the system to stabilize from the starting stage of the simulation) is obtained from the:

$$F_{exc}(t) = \Re\left[\frac{H}{2} F_{exc}(\omega, \theta) e^{i\omega t}\right] \qquad (30)$$

where $\Re$ denotes the real part of the term in bracket, $H$ is the wave height, $F_{exc}$ is the frequency dependent complex wave-excitation amplitude vector, and $\theta$ is the wave direction. The excitation force in irregular sea states can be calculated as follows:

$$F_{exc}(t) = \Re\left[\sum_{j=1}^{N} F_{exc}(\omega_j, \theta) e^{i(\omega t + \phi)} \sqrt{2S(\omega_j) d\omega_j}\right] \qquad (31)$$

where *N* is the number of frequency bands that discretizes the wave spectrum, φ is the randomized phase angle, and $S(\omega)$ is the distribution of wave energy over a range of wave frequencies that are characterized by a $H_s$ and $T_p$.

The software uses the following equation to calculate the radiation terms, namely the added mass and the radiationdamping torques. In this equation, the first term is the added mass torque, and the second term is the radiation-damping torque:

$$F_{rad}(t) = -A_\infty \ddot{X} - \int_0^t K_r(t-\tau)\dot{X}(\tau)d\tau \qquad (32)$$

where $A_\infty$ is the added mass matrix at an infinite frequency, and $K_r$ is the radiation impulse response function, which is calculated by this equation:

$$K_r t = \frac{2}{\pi}\int_0^\infty B(\omega)\cos(\omega t)d\omega \qquad (33)$$

Notably, the assumption is that there is no motion before t = 0. The $A_\infty$ and $B(\omega)$ coefficients are calculated by the NEMOH, the BEM solver WEC-Sim uses.

*5.3. Optimization Run Details*

In order to optimize the performance of an OSWEC in the southern Caspian Sea, WEC-Sim was used to simulate the converter and HC-EGWO to optimize its power output. A simulation time of 400 s and a ramp time of 100 s were chosen, with time steps of 0.1 s. Furthermore, ten optimization runs using HC-EGWO were performed, each with 1000 iterations and 20 search agents.

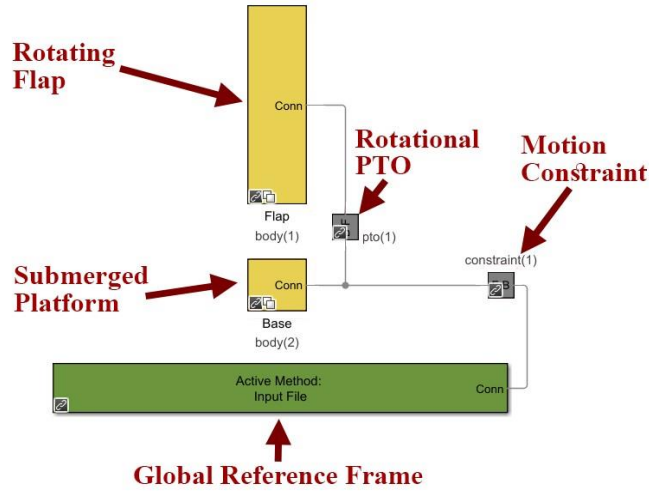

Figure 13: OSWEC's Simulink model in WEC-Sim [23]

Table 4: The details of optimisation parameters and other variables related to the wave power simulation applied.

| Abbreviation | Full name | Description |
|---|---|---|
| H | Height (m) | Measure of the amplitude or intensity of a wave |
| T | Period (s) | Time for completing one full cycle of wave |
| K | PTO Stiffness Coefficient (MNm/rad) | Relationship between the deformation of a PTO system and the force it generates |
| C | PTO Damping Coefficient (MNsm/rad) | Relationship between the PTO system's velocity and the force it generates |
| FlapEP | maxFlapExcitationPitch | Maximum Flap's Excitation Force (kN) |
| FlapRDP | maxFlapRDPitch | Maximum Flap's Radiation Damping Force (kN) |
| FlapAMP | maxFlapAMPitch | Maximum Flap's Added Math Force (kN) |
| FlapRP | maxFlapRestoringPitch | Maximum Flap's Restoring Force (kN) |
| ForceTPTOP | maxForceTotalPTOPitch | Maximum PTO's Force (kN) |
| meanFlapAVD | meanFlapAngularVelocityD | Average Flap's Velocity (degree/s) |
| maxFlapAVD | maxFlapAngularVelocityD | Maximum Flap's Velocity (degree/s) |
| FlapARD | maxFlapAngularRotationD | Maximum Flap's Rotation (degree) |
| Flapa | flap's acceleration (degree/$s^2$) | Affects the PTO system's structural integrity & overall performance. |
| Flapfam | flap's force added math (N) | |
| Flapfe | flap's excitation force (N) | A function of the displacement, velocity and acceleration of PTO system |
| Flapfr | flap's restoring force (N) | Result of the buoyancy and gravity forces acting on the WEC |
| Flapfrd | flap's radiation damping force (N) | Due to the interaction between a WEC and the surrounding water waves |
| Flapft | flap's total force (N) | Sum of hydrodynamic forces, gravity forces, buoyancy forces, and other forces generated by the PTO system. |
| Flapv | flap's velocity (degree/s) | Rate of change of the flap angle, |
| Flapx | flap's position (degree) | |
| PTOa | PTO's acceleration (degree/$s^2$) | Alteration rate of its velocity over time |
| PTOv | PTO's velocity (degree/s) | Alteration rate of its position over time |
| PTOf | PTO's force (N) | Is transmitted from the WEC to the PTO system due to the motion of the waves |
| PTOx | PTO's position (degree) | |

## 6. Results

### 6.1. Benchmark Functions Results

Based on the results of the landscape analysis, it has been revealed that the problem at hand exhibits a multimodal nature. Therefore, multimodal benchmark functions have been used to test the effectiveness of HC-EGWO to optimize a wide range of complex problems. Furthermore, assessing the performance of HC-EGWO using multimodal benchmark functions can help in clarifying the generalization ability of the optimisation method. The importance of generalization lies in its ability to prevent overfitting [67], a situation where an optimization algorithm excessively fine-tunes its parameters to match specific conditions perfectly. Overfitting can result in subpar performance when the algorithm is applied to unfamiliar problem instances. By giving priority to generalization, optimization methods concentrate on capturing fundamental patterns and principles that can be transferred to new problem instances, resulting in solutions that are more dependable and efficient.

These benchmark functions have multiple global and local optima, which increase by the number of dimensions. This characteristic makes them the perfect functions to test the exploration ability of an algorithm [58]. As shown in Table 5, the HC-EGWO can provide very competitive results (especially in fixed-dimension multimodal benchmark function). This algorithm reaches the best solutions in 7 test functions and the second-best answer in 2 functions in this category, which is the best performance among the analyzed algorithms. It is notable that in some functions, like F11 and F12, the difference in performance is very minuscule.

Table 5: Results of the multimodal benchmark functions.

| F | GWO Avg. | Std. | mGWO Avg. | Std. | EEGWO Avg. | Std. | IGWO Avg. | Std. | ERGWO Avg. | Std. | HC-EGWO Avg. | Std. |
|---|---|---|---|---|---|---|---|---|---|---|---|---|
| 1 | -5.792E+03 | 6.756E+02 | -5.347E+03 | 1.046E+03 | -2.096E+03 | 5.089E+02 | -6.054E+03 | 8.322E+02 | -2.202E+03 | 4.659E+02 | -5.743E+03 | 8.293E+02 |
| 2 | 3.177E+00 | 3.672E+00 | 2.842E-14 | 3.580E-14 | 0.000E+00 | 0.000E+00 | 7.161E+00 | 4.462E+00 | 0.000E+00 | 0.000E+00 | 1.074E+02 | 3.002E+01 |
| 3 | 1.007E-13 | 1.927E-14 | 2.164E-14 | 4.324E-15 | 4.441E-16 | 0.000E+00 | 1.977E-09 | 1.238E-09 | 3.997E-15 | 0.000E+00 | 5.184E-14 | 1.594E-14 |
| 4 | 4.783E-03 | 7.820E-03 | 1.508E-03 | 4.962E-03 | 0.000E+00 | 0.000E+00 | 9.001E-03 | 1.102E-02 | 0.000E+00 | 0.000E+00 | 1.718E-03 | 5.688E-03 |
| 5 | 3.634E-02 | 1.601E-02 | 3.843E-02 | 1.752E-02 | 1.431E+00 | 1.188E-01 | 7.412E-02 | 5.245E-02 | 8.409E-01 | 1.875E-01 | 5.903E-07 | 3.399E-07 |
| 6 | 5.820E-01 | 2.346E-01 | 5.388E-01 | 2.066E-01 | 2.995E-03 | 1.932E-03 | 8.668E-01 | 2.751E-01 | 2.970E+00 | 4.908E-02 | 1.318E-02 | 3.027E-02 |
| 7 | 5.492E+00 | 4.853E+00 | 5.754E+00 | 4.788E+00 | 1.161E+01 | 2.764E+00 | 5.628E+00 | 4.552E+00 | 9.107E+00 | 3.433E+00 | 4.721E+00 | 4.109E+00 |
| 8 | 5.662E-03 | 1.223E-02 | 5.664E-03 | 1.220E-02 | 1.795E-02 | 1.400E-02 | 3.733E-03 | 7.566E-03 | 2.822E-03 | 2.957E-03 | 2.482E-03 | 6.065E-03 |
| 9 | -1.032E+00 | 1.641E-08 | -1.032E+00 | 7.774E-08 | -5.038E-01 | 2.333E-01 | -1.032E+00 | 2.881E-12 | -1.018E+00 | 1.218E-02 | -1.032E+00 | 5.206E-12 |
| 10 | 3.979E-01 | 5.884E-04 | 3.979E-01 | 3.995E-05 | 2.550E-05 | 1.985E+00 | 3.979E-01 | 7.629E-11 | 7.928E-01 | 3.432E-01 | 3.980E-01 | 4.306E-04 |
| 11 | 3.000E+00 | 4.631E-05 | 3.000E+00 | 2.031E-05 | 6.862E+01 | 7.135E+01 | 3.000E+00 | 6.642E-05 | 1.012E+01 | 1.037E+01 | 3.000E+00 | 2.381E-05 |
| 12 | -3.861E+00 | 2.341E-03 | -3.862E+00 | 2.260E-03 | -3.242E+00 | 4.768E-01 | -3.862E+00 | 2.581E-03 | -3.424E+00 | 3.243E-01 | -3.862E+00 | 1.933E-03 |
| 13 | -3.279E+00 | 6.838E-02 | -3.281E+00 | 6.435E-02 | -1.601E+00 | 5.947E-01 | -3.278E+00 | 7.025E-02 | -2.003E+00 | 4.288E-01 | -3.241E+00 | 7.815E-02 |
| 14 | -8.805E+00 | 2.541E+00 | -9.404E+00 | 1.949E+00 | -7.846E-01 | 1.940E-01 | -9.643E+00 | 1.556E+00 | -2.461E+00 | 5.867E-01 | -9.205E+00 | 2.190E+00 |
| 15 | -1.022E+01 | 9.702E-01 | -1.022E+01 | 9.694E-01 | -8.125E-01 | 2.101E-01 | -1.023E+01 | 9.704E-01 | -2.383E+00 | 7.936E-01 | -1.005E+01 | 1.343E+00 |
| 16 | -9.814E+00 | 2.238E+00 | -1.008E+01 | 1.750E+00 | -9.005E-01 | 2.442E-01 | -9.790E+00 | 2.326E+00 | -2.470E+00 | 5.635E-01 | -1.054E+01 | 6.584E-08 |

Figure 14 shows a comparative plot of the five variants of GWO and the proposed GWO (HC-EGWO) performance over the 16 benchmarks. The performance average rank of each variant and the significant differences using the Friedman test can confirm that HC-EGWO performed best in these 16 multi-modal optimisation benchmarks. The average rank is used to calculate the Friedman statistic, which is compared to the critical value to determine whether there are significant differences in performance.

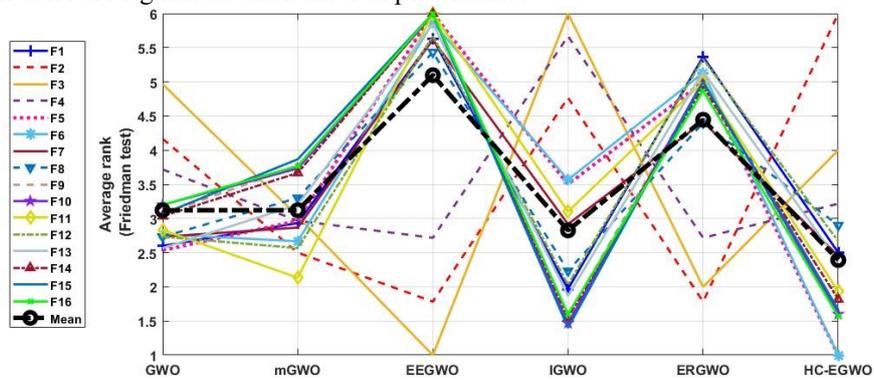

Figure 14: Average performance rank of the HC-EGWO compared with other five variants of GWOs over the 16 benchmarks using the Friedman test.

*6.2. Algorithms Performance in the Defined Problem*

Here, the proposed algorithm was compared to the conventional GWO and its four variants that were introduced earlier. Each algorithm was run five times with a population size of 20 and 1000 iterations to achieve maximum power output. Table 6 shows the critical parameters of the OSWEC in the average performance of each algorithm.

As can be seen in Table 6, all the algorithms have competitive performances; however, the proposed algorithm (HC-EGWO) outperforms the others. HC-EGWO can improve the power output by 0.08% up to 3.31% compared to the other candidates. Moreover, the EEGWO has the worst performance by far. Next, the convergence curves for the six inspected algorithms are presented in Figure 15.

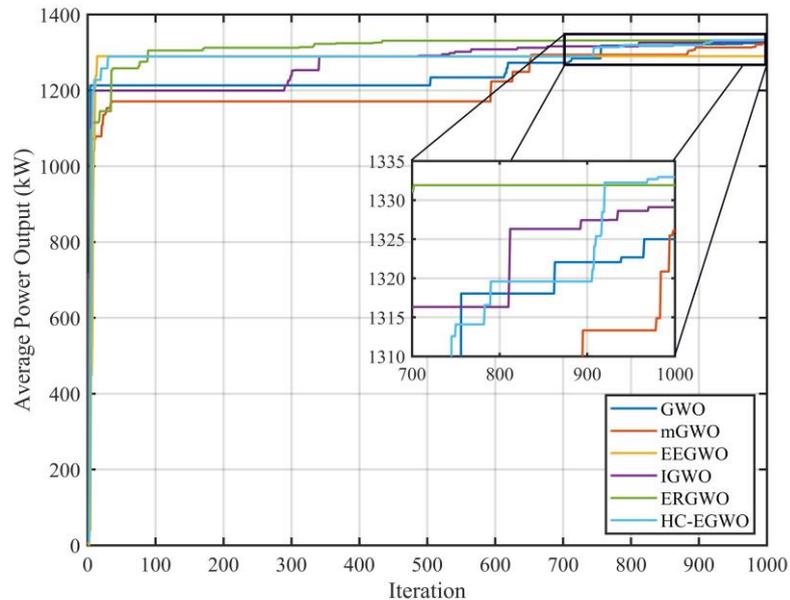

Figure 15: The convergence curves for the average performance of the conventional GWO, its four modifications, and the proposed algorithm in this study.

Figure 15 shows how competitive all these algorithms' performances were in this problem. In addition, all the methods were able to reach high amounts of average power output in under 50 iterations.

*6.3. Wave and PTO parameters optimization*

In this section, the results of the wave and PTO parameters optimization of the oscillating wave surge converter are presented. In order to get a better understanding of the effects of optimization on performance, the converter's functioning and outputs in 3 scenarios are analyzed and compared. One of the cases is the scenario with the best-found solution by the HC-EGWO (Case C). Next, the case with the default WEC-Sim parameters was chosen to see how much improvement the input fine-tuning has achieved (Case A). However, since the default WEC-Sim case parameters were in the unfeasible area following the literature review this study performed at the research's beginning, another scenario was added for evaluation. It was observed that the PTO damping with a value of 0.012 MNsm/rad was in the unfeasible area, so based on the literature review and the initial sensitivity analysis, the minimum feasible value, which was 90 MNsm/rad, was chosen for the following case, and the three other parameters stayed the same (Case B).

Table 7 presents the inputs, forces, oscillation, and power of the system in the three analyzed cases in detail. First, the wave elevation during the simulation for the 3 cases is presented in Figure 16 in order to analyze the other parameters more effectively.

Next, the resulting oscillation details, PTO force, and power output will be inspected. As stated before, cases A and B have the same wave conditions. Hence, one wave elevation graph is plotted to represent the sea state in both cases in Figure 16. According to this figure, in cases A and B, wave elevation relatively stays in the same range, and the amplitude does not change drastically at any point during the simulation. On the other hand, in case C, especially after the halfway mark, wave heights are greater and reach their maximum absolute value at around the 215-second mark. As previously mentioned, when using linear PTO for the OSWEC, WEC-Sim calculates the power output by multiplying the PTO force by the flap's angular velocity. Both the flap's motion and its angular velocity positively dictate the device's power output, which is this study's main objective.

Table 7: The details of the 3 analyzed cases in this study

| | H | T | K | C | FlapEP | FlapRDP | FlapAMP |
|---|---|---|---|---|---|---|---|
| **Case A** | 2.5 | 8 | 0 | 0.012 | 1624 | 1778 | 691.7 |
| | FlapRP | ForceTPTOP | meanFlapAVD | maxFlapAVD | FlapARD | MaxPower | AvgPower |
| | 887.4 | 4.158 | 7.049 | 19.85 | 26.29 | 1.440 | 0.228 |
| **Case B** | H | T | K | C | FlapEP | FlapRDP | FlapAMP |
| | 2.5 | 8 | 0 | 90 | 1624 | 780.3 | 297.6 |
| | FlapRP | ForceTPTOP | meanFlapAVD | maxFlapAVD | FlapARD | MaxPower | AvgPower |
| | 360.6 | 13395 | 2.733 | 8.527 | 10.686 | 1993 | 287.7 |
| **Case C** | H | T | K | C | FlapEP | FlapRDP | FlapAMP |
| | 4.223 | 7.39 | 54.46 | 75.58 | 3687 | 1512 | 722 |
| | FlapRP | ForceTPTOP | meanFlapAVD | maxFlapAVD | FlapARD | MaxPower | AvgPower |
| | 842 | 38306 | 3.82 | 21.92 | 24.97 | 13393.98 | 1332.95 |

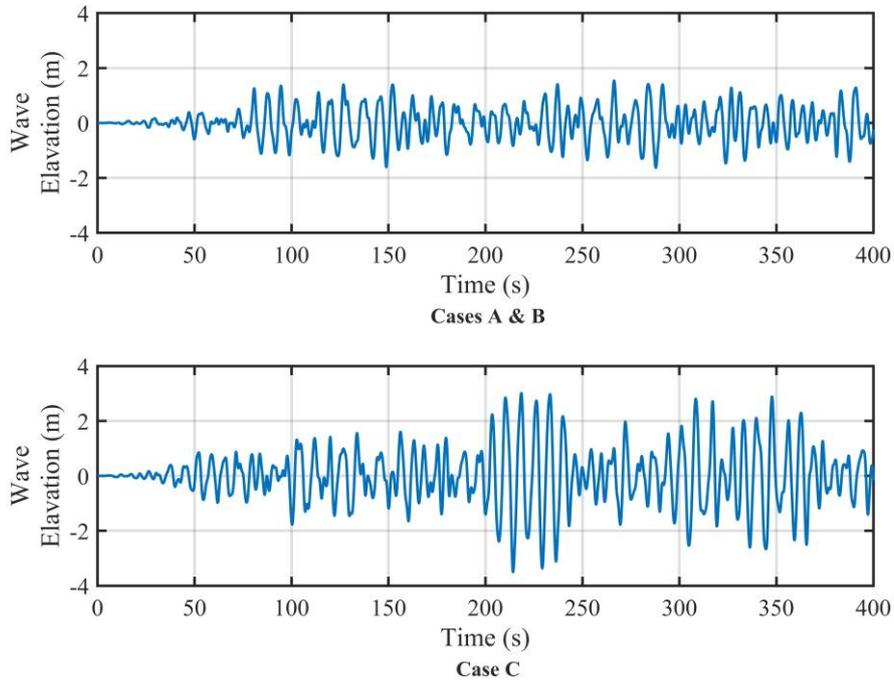

Figure 16: Wave elevation profile during the performance of the WEC for the 3 cases

First, we compare the flap's oscillation in cases A and B in Figure 17. Since these two have the same wave characteristics but different PTO configurations, one can assign almost all the difference in oscillation to the PTO stiffness and damping. Both parameters are virtually trivial in case A and come into effect in case B. It can be seen that the PTO C and K in isolation dampen the flap's oscillations, both the motion and the velocity. For case C, the flap's fluctuation during the simulation almost mimics the shape of the wave elevation, which is predictable. But the maximum flap motion roughly occurs in t = 125 s of case A.

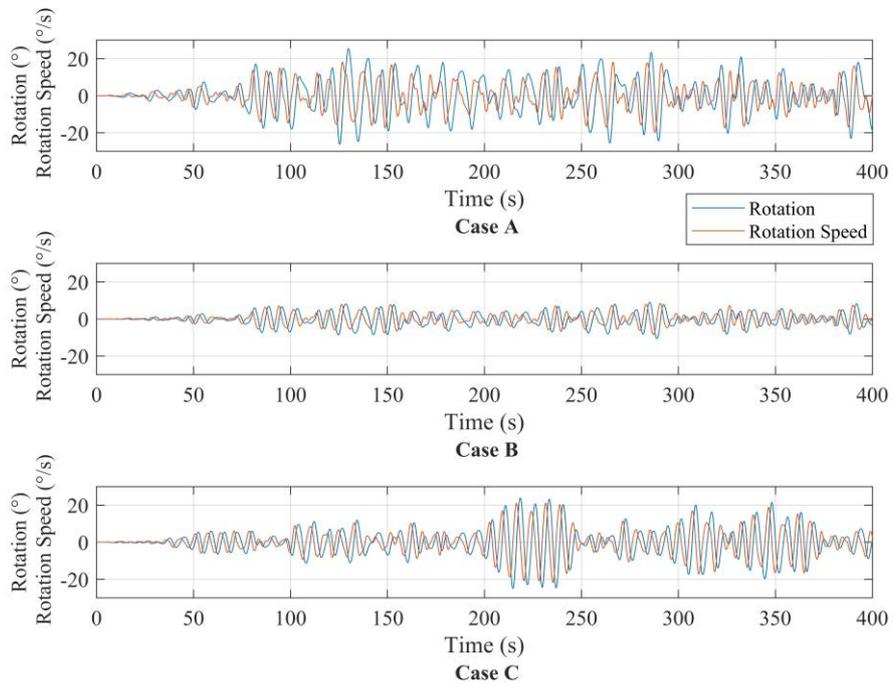

Figure 17: Flap's oscillations motion and velocity the 3 cases

Next, in Figure 18 the PTO force and power output for the 3 cases are presented. Note that for this purpose, the y-axis for the 3 cases is modified for more clear visualization. According to the y-axis, it can be said that roughly case B produces ten times more power the case A produces. And that case C generates 50 times more the case A.

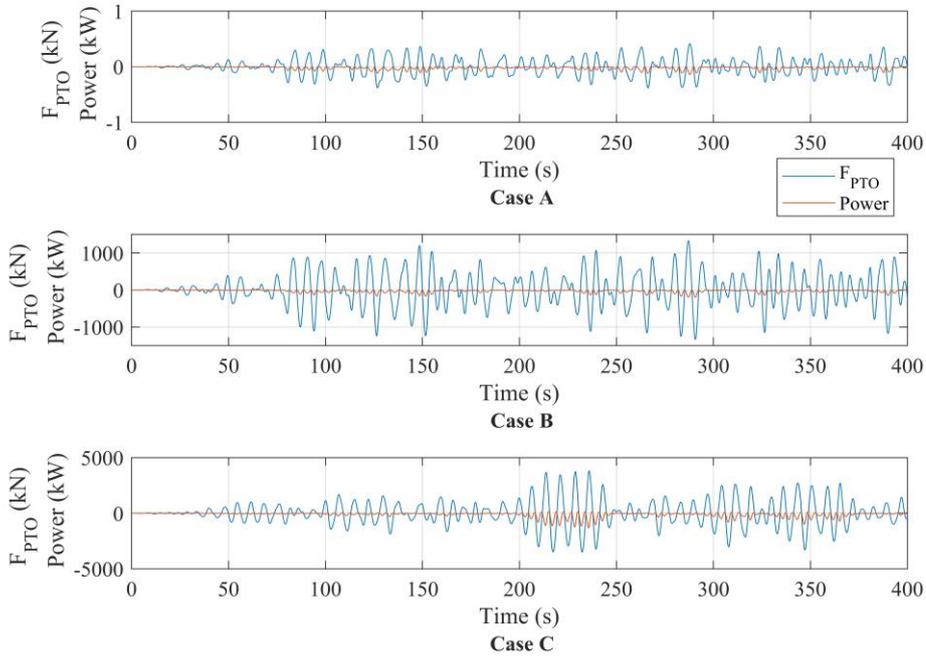

Figure 18: PTO force and power output for the 3 cases

For case C, the PTO force is peaking during the 200 s and 250 s marks due to the substantial wave magnitude (see Figure 16). This happens multiple times to a lesser extent at different times during the simulation (100-140 s, 155-165 s, 175-180 s, 270-375 s). Similarly, it can be seen for case B that the highest power outputs occur when the PTO force is at max. But overall, the extent of the produced power is about five times smaller, which can be attributed to the PTO mechanical parameters. Finally, for case A, the default WEC-Sim case, considering the PTO stiffness coefficient is zero and the damping coefficient is very low, the generated power is almost between 500 and 1000 times smaller than the best-found sample in case C. In the context of optimizing the power output of a WEC, parallel coordinate plots can be helpful in understanding the relationships between the optimization parameters (such as damping, stiffness, wave height, and period) and the resulting power output. For example, in Figure 19, we can observe that increasing the damping and stiffness of the WEC leads to a decline in power output, while increasing the wave height leads to an upsurge in power output.

Figure 19 showcases the parallel plots for the two selected runs of the HC-EGWO in order to achieve the highest power output; this includes the best-found solution by all the ten optimization runs. Furthermore, by analyzing the lines corresponding to each parameter in this Figure, it can be achievable to identify the range of parameter values that lead to optimal power output; for instance, the optimal ranges of K and C are [50-65] and [70-80], respectively. Another significant observation from the parallel plots is that there are sharp, non-linear relationships between the optimization parameters and the power output of the WEC.

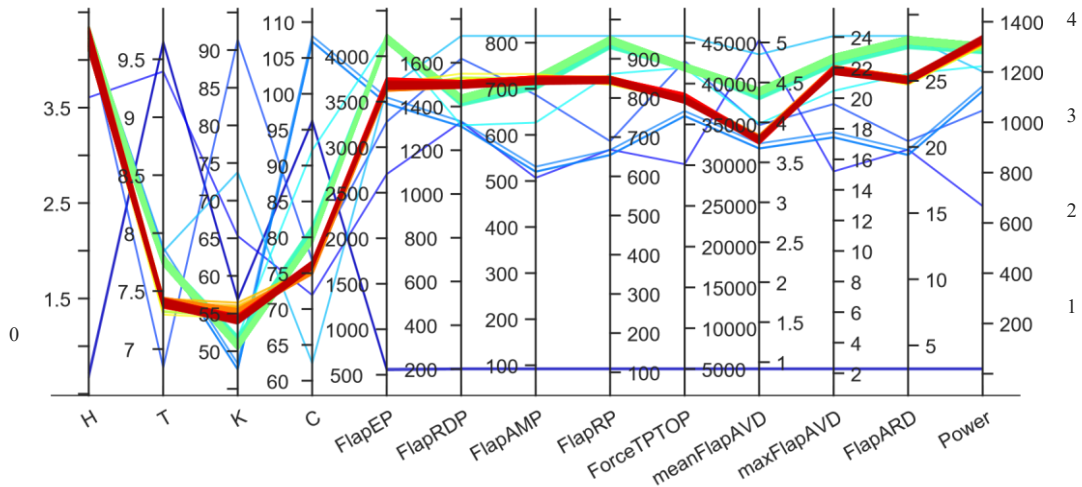

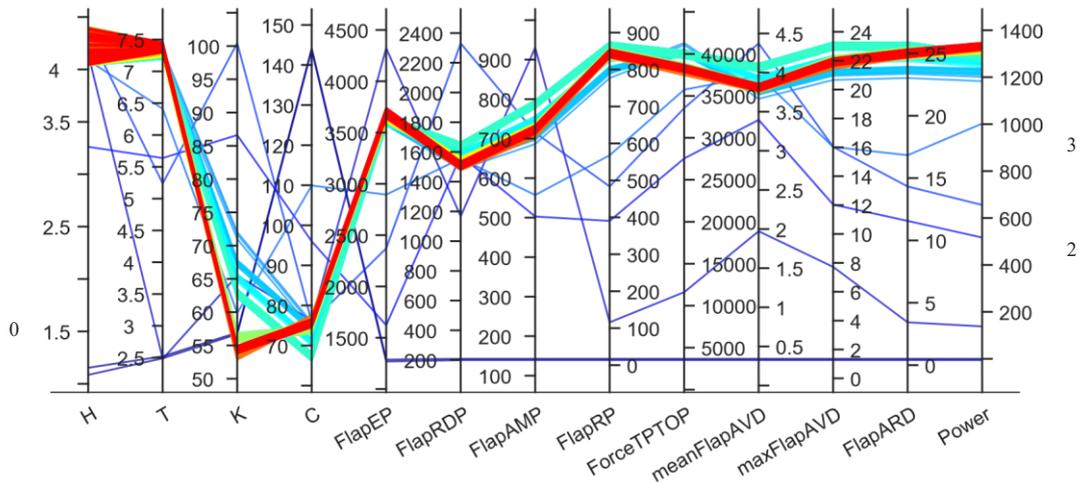

Figure 19: Two examples of the best-performed optimization method's exploration through the decision variables (H, T, K, and C) with internal parameters of the simulator listed in Table 4, plus the average of total power output distribution visualized by a parallel coordinates plot. The dark red lines indicate the highest absorbed power output based on the configurations. (in the figures, the H, T, K, and C are the input parameters. FlapEP, FlapRD, and FlapAMP are, respectively, the excitation, radiation damping, and added mass torques, and the ForceTPTOP is the PTO force. The meanFlapAVD and maxFlapAVD are the average and maximum angular velocity of the flap, the FlapARD is the flap's angular rotation of the flap, respectively, and the Power is the average power output of the system.)

Figure 20 shows the parallel plots for the three scenarios studied in the result (See Section 6.3). These data are in real-time during the simulation time. Next, all the y-axis, except for the power output, are symmetrical, showcasing the device's oscillating nature and, therefore, its parameters. But when taking a closer look at case A (Figure 20(a)), it is visible that the only parameters that have the simultaneous maximum as the power output (red lines) are the flap's angular velocity and PTO force, this is consistent with equations for calculating the power output in WEC-Sim in earlier sections of the study. And also corresponds to the moderate values of the flap's acceleration, restoring torque, excitation torque, and other hydrodynamic forces. Based on case C (Figure 20(c)), it can be seen that low absolute values of excitation force correspond to low power. Since both these parameters positively correlate with wave height, that is consistent with the theory. Notably, not every hydrodynamic force has the same trajectory as the power; for instance, the restoring force's maximum absolute values coincide with the lowest power outputs. Since the flap's restoring torque is dependent on the flap's displacement only, we can see that the displacement alone can not lead to the best performance. The flap's velocity can be considered a more critical and deciding factor.

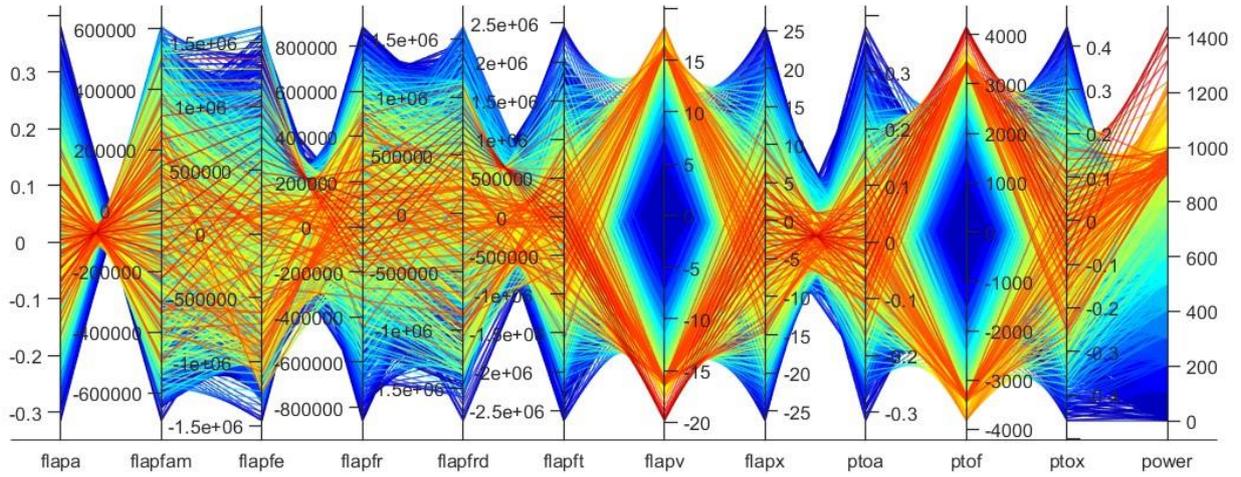

(a)

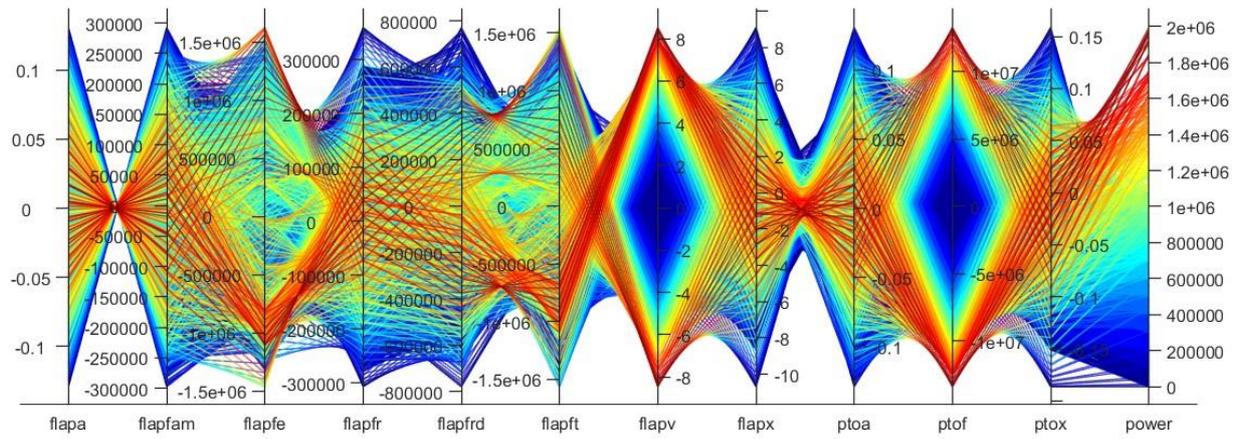

(b)

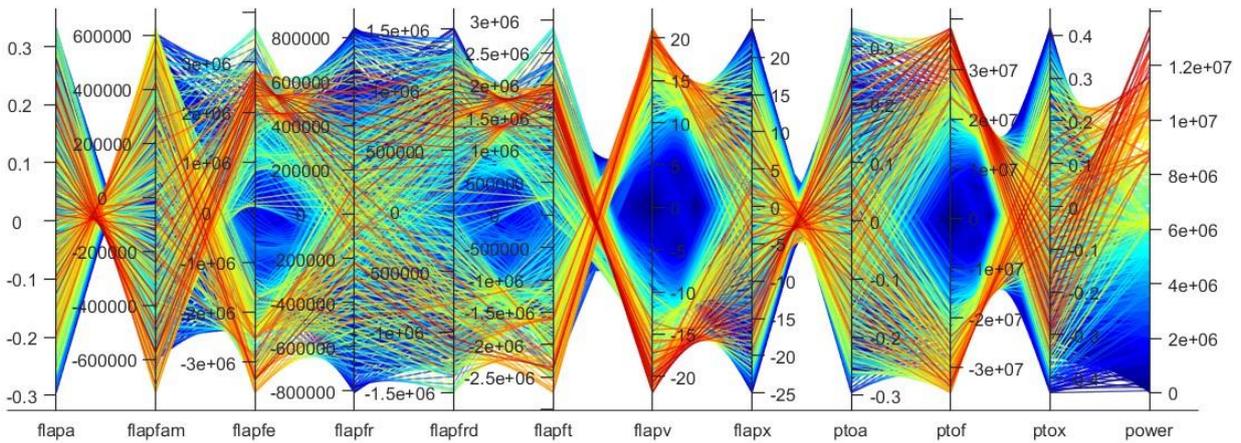

(c)

Figure 20: Parallel interactions plot for the average of total power output distribution based on three scenarios described in the Result section. The technical details of the variables are listed in Table 4.

*6.4. Sensitivity Analysis*

Sensitivity analysis is a crucial mechanism in post-processing optimization methods due to identifying the most significant factors affecting the efficiency of the optimized models [68]. Table 7-Case C reports the best configuration of decision variables (H, T, K, and C) and internal hydrodynamic parameters of the simulator proposed by the HCEGWO. Meanwhile, the sensitivity analysis results can be seen in Figure 21. The black lines show the unfeasible areas of the search space for K (Figure 21(c)) and C (Figure 21(d)). The power output of the best-found solution discovered by sensitivity analysis was 1333.1 (kW). This negligible improvement can confirm that the proposed optimization method (HC-EGWO) is able to explore comprehensively the search space and converge to an appropriate solution.

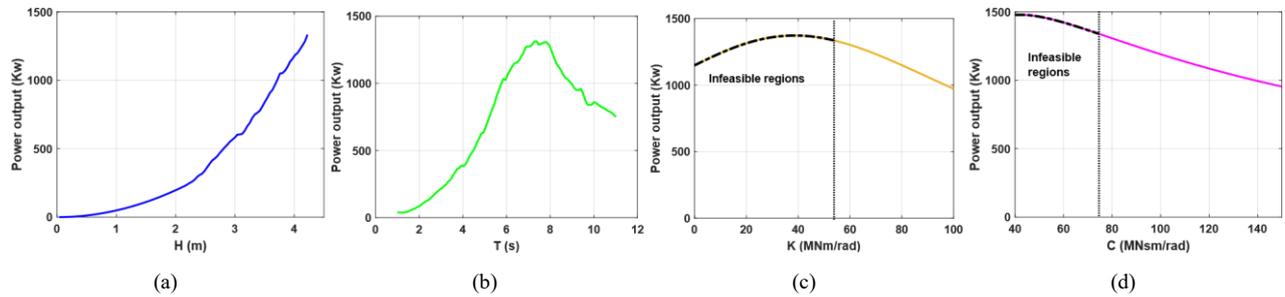

Figure 21: Sensitivity analysis of the best-found configuration using the proposed hybrid optimisation method.

*6.5. Site Selection*

In the last segment, an operational spot will be selected as the best location for the installation of the device. This is obtained by analyzing the best-found solution and finding the location from the 105 initial data points in the Caspian Sea that has the closest wave characteristic values, namely wave height and wave period, to the theoretic best location. The optimum values are H = 4.223 m and T = 7.39 s. Then, the best location was found using a Root Mean Square Error method (Figure 22), and the RMSE values for all the data points were evaluated. In the end, a data point belonging to the Kiashahr Port resulted in RMSE = 2.78, which was the minimum among the analyzed spots. The longitude and latitude of the best-found location are 37.6° N 50.1° E; this spot belongs to the Kiashahr Port.

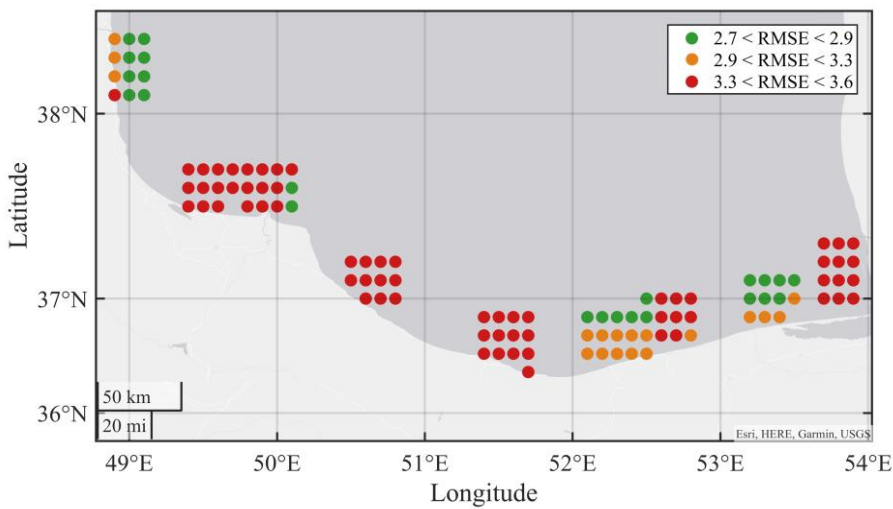

Figure 22: Categorization of the 105 data points based on the fitness of their wave significant height and peak period (using the RMSE method)

7. Conclusion

This study focused on optimizing the power take-off (PTO) parameters and site selection for an offshore oscillating surge wave energy converter (OSWEC) in the Caspian Sea. The optimization was performed using the novel Hill Climbing Explorative Gray Wolf Optimizer (HC-EGWO), which showed strong performance across 16 benchmark multimodal functions. When applied to the OSWEC case study, the HC-EGWO discovered a high-quality solution that increased the power output by approximately 3% compared to other methods.

The results provide valuable insights into the complex interplay between the converter's mechanical design and the surrounding wave climate. Specifically, the sensitivity analyses revealed that wave height and PTO damping have the most substantial impact on the power output. Increasing the wave height boosts the absorbed power, while lower PTO damping is preferable. Meanwhile, moderate values of wave period and PTO stiffness lead to the highest outputs. There are also non-linear relationships and trade-offs between the parameters influencing the hydrodynamic forces acting on the device.

Overall, the proposed HC-EGWO algorithm proved effective in handling this challenging multimodal optimization problem. The hybridization with local search prevented premature convergence and bolstered the exploration of the solution space. The outcomes showcase the method's capabilities for optimizing offshore renewable energy systems where complex hydrodynamic interactions are involved. They provide a valuable starting point for devising control strategies that ensure OSWECs operate safely within extreme seas while maximizing power generation.

Moreover, the results offer insights into deploying OSWECs in the unique conditions of the Caspian Sea. The landscape analysis of available wave data from the region informed the creation of feasible parameter bounds. And the site selection analysis pinpointed a location with strong energy potential based on the optimal wave height and period found by the HC-EGWO. Hence, the outcomes provide a launchpad for harnessing the vast untapped wave resources of the Caspian basin.

Future work can focus on incorporating more advanced hydrodynamic modeling into the WEC simulations. The effects of viscosity, turbulence, and non-linear waves could be accounted for using computational fluid dynamics. Additionally, real-world PTO systems like hydraulic and direct-drive PTOs can be simulated to optimize their parameters. Finally, expanding the optimization to more complex problems like WEC arrays and combining it with control strategy optimization represents promising research directions.